\documentclass[conference]{IEEEtran}
\IEEEoverridecommandlockouts
\usepackage{cite}
\usepackage{amsmath,amssymb,amsfonts}
\usepackage{graphicx}
\usepackage{textcomp}
\usepackage{xcolor}
\usepackage{textcomp}
\usepackage{amsmath}
\usepackage{soul}
\usepackage{color}
\usepackage{makecell}
\usepackage{booktabs}
\usepackage{lipsum}
\usepackage{tikz}
\usetikzlibrary{arrows.meta,positioning,fit}
\usepackage{algpseudocode}
\usepackage{stfloats}

\def\BibTeX{{\rm B\kern-.05em{\sc i\kern-.025em b}\kern-.08em
    T\kern-.1667em\lower.7ex\hbox{E}\kern-.125emX}}
\begin{document}

\title{From Prompt to Service: An SLM-Based Agent Orchestration Gateway for AI-Driven Virtual Worlds\\}

\author{\IEEEauthorblockN{Louis Nisiotis}
\IEEEauthorblockA{\textit{Department of Computing, Engineering and Mathematics} \\
\textit{UCLan Cyprus} 
\\
LNisiotis@uclan.ac.uk}

\and
\IEEEauthorblockN{Aimilios Hadjiliasi}
\IEEEauthorblockA{\textit{Department of Computing, Engineering and Mathematics} \\
\textit{UCLan Cyprus} 
\\
AHadjiliasi1@uclan.ac.uk}
}
\maketitle

\begin{abstract} 
AI-driven virtual worlds depend on heterogeneous AI services with different computational requirements and deployment locations. 
Directly embedding these capabilities in the client reduces extensibility and complicates coordination across edge and cloud infrastructure.
This paper presents an \textit{SLM-based Agent Orchestration Gateway}, a lightweight runtime coordination mechanism that decouples a virtual world client from heterogeneous AI backends through intent-driven service routing. 
An edge-deployed SLM classifies the semantic intent of each user prompt, a configurable service registry validates and resolves the routing decision, and the selected backend is invoked, enabling new AI capabilities to be introduced in the virtual world without modifying the client application. 
The gateway is implemented and evaluated within the \textit{InterwovenXR} virtual museum testbed. 
The evaluation shows that compact SLMs can provide reliable service selection based on intent recognition capabilities on edge hardware, and that task-specific fine-tuning can transform sub-billion-parameter models into practical, low-latency routers. 
A layered configuration pairing a fine-tuned sub-billion-parameter model as router with a larger SLM for conversational response generation is shown to be deployable on mid-range edge hardware and more efficient than delegating both responsibilities to a single model.
The findings show that SLMs can support practical AI service orchestration in virtual worlds and the work contributes an evaluated architecture for scalable, extensible, and edge-supported AI interaction, enabling virtual agents become access points to distributed generative AI services.

\end{abstract}

\begin{IEEEkeywords}
Small Language Models, Agent Orchestration, Virtual Worlds, Edge AI, Generative AI, Virtual Agents

\end{IEEEkeywords}

\section{Introduction}
Advances in eXtended Reality (XR), real-time rendering, distributed infrastructures, and interoperability standards have expanded the technical foundations of virtual worlds, enabling immersive interaction, persistency, cloud-edge execution, and cross-platform deployment for enriched interactive experiences \cite{lyu2024state, yang2025interoperability}.
Within this context, a vast amount of recent research in S/LLM-powered virtual agents and generative AI systems is explored as runtime components of immersive environments.
Examples include conversational avatars that support dialogue, assistance, and adaptive interaction \cite{wang2024agents,nisiotis2025compsac}, LLMs for prompting interactive worlds,  generating 3D objects, procedural creation of virtual elements from language prompts and other \cite{delatorre2024llmr, siddiqui2024meshgpt,jiang2024survey}.
In fact, the integration of generative AI  into XR environments for immersive experiences is changing the technical design of virtual worlds. 
Emerging AI-driven virtual worlds need to coordinate specialised services at runtime where user requests may require lightweight conversational response generation, translation, accessibility adaptation, retrieval from a domain knowledge base, 3D asset generation from a text-to-3D backend and many other, and are key capabilities for modern XR simulations, collaborative workspaces, and intelligent interactive applications \cite{nisiotis2025itpro,Chamola2024,bengesi2024advancements,wang2024agents}. 
This shift creates a runtime orchestration problem where directly coupling the virtual-world client to each service, endpoint, payload, and deployment location reduces extensibility and complicates routing, validation, latency management, and fallback handling.
The problem becomes more important when services are distributed across local machines, edge devices, institutional servers, and cloud-based generative AI systems. 
In such settings, the virtual world requires a coordination layer that can interpret user intent, select the correct service, and return a response in a format that the environment can process.
Prior work has examined generative AI in virtual worlds through conversational agents, procedural content generation, and text-to-3D systems \cite{delatorre2024llmr,siddiqui2024meshgpt,jiang2024survey,nisiotis2025compsac}, but less attention has been given to coordinating heterogeneous AI services during live interactions.
This paper addresses that gap by presenting an SLM-based Agent Orchestration Gateway that operates between a virtual world client and multiple AI backends, using an edge-deployed orchestration mechanism utilising SLM as an intent router for service selection.
The contribution of this work lies in the design, implementation, and empirical evaluation of an intent-driven orchestration system for heterogeneous AI services in interactive virtual worlds. 
Specifically, the work contributes: 1) an Agent Orchestration Gateway that decouples the virtual world client from heterogeneous AI backends through a configurable service registry and semantic intent routing; 
2) an empirical evaluation of SLMs as edge-deployed service routers, quantifying routing accuracy, structured-output reliability, and latency; 
and 3) an evaluation of a two-model edge configuration in which a fine-tuned sub-billion-parameter model performs low-latency service selection and a larger SLM is activated for conversational response generation.

\section{Background and Context}
\subsection{Immersive Virtual Worlds}
Virtual worlds have evolved from persistent multi-user environments into complex interactive systems featuring avatar-mediated communication, persistent shared spaces, user-generated content, XR interfaces, real-time networking, spatial audio, interactive objects, and intelligent agents among other features \cite{dionisio2013virtual,mystakidis2022metaverse,nisiotis2025itpro}. 
In fact, virtual worlds have found their way into real life contexts beyond socialisation and entertainment, for example in education with immersive simulation, experiential learning, virtual tutoring, and collaborative activities \cite{wang2022edumetaverse,chen2023metaverse,hadjiliasi2025eduverse},  industrial contexts with digital twins, simulations, remote maintenance, training, and cyber-physical monitoring \cite{siyaev2021aircraft,wang2023digitaltwins,njoku2023transportation}, professional workspace with meetings and virtual presentations and other, highlighting their relevance and appeal for immersive interactions.
The technical maturity of virtual worlds has been strongly influenced by advances in XR hardware, real-time computer graphics, rendering, and GPU acceleration. 
Modern software development tools and game engines such as Unity and Unreal Engine provide mature tools for user interactions, gameplay logic development, physics, animation, lighting, networking, and cross-platform deployment, enabling the construction of high-fidelity virtual environments with interactive behaviours and support cross platform deployment.
The development ecosystem has also improved through interoperability standards such as OpenXR, WebXR, glTF, and OpenUSD supporting device access, asset portability, and 3D scene exchange across heterogeneous platforms, reducing fragmentation in virtual world development, interaction data and platform-specific constraints \cite{yang2025interoperability}.
Interaction modalities have also expanded beyond keyboard, mouse, and controller input as were commonly used in interactive 3D environments.
Modern virtual worlds integrate gaze tracking, hand tracking, gesture recognition, spatial audio, haptic feedback, motion capture, wearable sensors, and physiological sensing to support in world interactions, improving embodiment and presence, supporting richer visual fidelity, lower development barriers, and more scalable deployment across educational, cultural, industrial, and entertainment scenarios \cite{nisiotis2025itpro,yang2025interoperability, kyrlitsias2022social}.

With the interest in deploying large-scale virtual environments and distributed systems leveraging edge and cloud network infrastructures increasing, there is a shift in reshaping how virtual worlds are deployed and experienced \cite{BelcastroCloudEdge, huponttorres2023}. 
Cloud services support large-scale persistence, shared state, asset streaming, and multi-user coordination, and edge computing brings computation closer to users and devices, reducing latency and limiting dependency on distant cloud infrastructure \cite{singh2023edgeai,qu2025mobileedge}. 
This is particularly relevant for AI-driven virtual worlds, where conversational agents, adaptive environments, and generated content must respond with minimal delay. 
Advances in 5G and emerging 6G networks further support these requirements by targeting higher bandwidth, lower latency, denser connectivity, and more reliable communication for distributed immersive systems \cite{adil2024metaverse5g6g}.

\subsection{Generative AI-Driven Virtual Environments}
Beyond hardware, software, and networking, the advent of generative AI has become a major driver of research interest in the next generation of virtual worlds.
Latest groundbreaking advancements in generative AI can produce text, images, audio, video, code, animation, and 3D assets from natural language or multimodal prompts \cite{chamola2024beyond,sengar2025generative}. 
Large Language Models (LLMs) and Small Language Models (SLMs) support dialogue, translation, summarisation, storytelling, planning, and reasoning, and text-to-image, text-to-audio, and text-to-3D systems provide new mechanisms for creating virtual content at runtime \cite{raiaan2024review,wang2024agents,siddiqui2024meshgpt} making virtual worlds intelligent adaptive systems where user intent can influence content, behaviour, and interaction flow.

Within virtual worlds, one of the most active application areas for generative AI is the development of conversational virtual agents. 
LLM-powered agents can provide natural dialogue, domain-specific guidance, multilingual support, adaptive tutoring, interactive storytelling, and contextual assistance within immersive spaces \cite{lau2024curling,jeon2023chatbots,hadjiliasi2025eduverse}. 
Compared with scripted agents, generative agents can respond more flexibly to user questions and support more open-ended forms of interaction. 
However, they also introduce challenges related to factual accuracy, hallucination, safety, latency, cost, privacy, and behavioural consistency, particularly in public-facing or educational environments \cite{huang2025hallucination,yao2024security,nisiotis2025isemv}.
These developments create the need for runtime architectures capable of coordinating heterogeneous AI services across local, edge, and cloud resources. 
A virtual world may require different AI capabilities depending on user intent, including lightweight conversational response generation, translation, domain-specific retrieval, specialist interpretation, or cloud-based 3D asset generation. 
Embedding each capability directly in the virtual world  can reduce extensibility and make it difficult to manage routing, safety, latency, monitoring, and service availability. 
The work presented in this paper responds to this architectural need through an SLM-based Agent Orchestration Gateway that operates between the virtual world and multiple generative AI services, enabling intent recognition, service routing, validation, and structured response delivery across edge and cloud backends.

\section{System Architecture and Virtual World}
To explore the possibilities offered by SLM-driven service orchestration in AI-supported virtual worlds, an Agent Orchestration Gateway has been designed and deployed within the \textit{InterwovenXR} virtual world testbed.
The gateway described in this section functions as a runtime coordination layer that receives natural language prompts from a virtual environment, performs intent recognition to identify the most appropriate AI service for each request, and returns a structured response, without requiring the client to maintain direct integrations with individual backend services. 
This design is particularly relevant to virtual worlds, where a single agent interface may need to handle conversational, linguistic, factual, specialist, generative and other requests, each potentially requiring a different model, computational resource, or deployment location.

\subsection{\textit{InterwovenXR} Virtual World Testbed}
The \textit{InterwovenXR} platform serves as the deployment environment for this work. 
It is a Cyber-Physical-Social virtual world testbed designed to support experimentation with XR interfaces, AI services, digital twins, robotic and virtual agents, and shared interactive environments \cite{nisiotis2026interwovenxr,nisiotis2020prototype,nisiotis2023interwoven,nisiotis2025compsac,nisiotis2025isemv}. 
The platform provides a shared digital space in which co-located users, remote participants, virtual agents, and cyber-physical assets interact with common content in real time, supporting research on human-agent communication, context-aware interaction, and AI-supported virtual-world experiences.
The use case of this study is a virtual museum focused on Cypriot cultural heritage, which includes a dedicated area for Panagia Aggeloktisti Church in Kiti, Cyprus. 
Visitors, represented as avatars, can navigate the environment, inspect digital reconstructions of the church architecture, apse mosaic, and iconostasi, and interact with virtual guide agents acting as curators. 
Multi-user communication is supported through Photon networking, text chat, VoIP, and instant messaging, enabling the museum to function as a shared social space.

\begin{figure}
    \centerline{\includegraphics[width=20.5 pc]{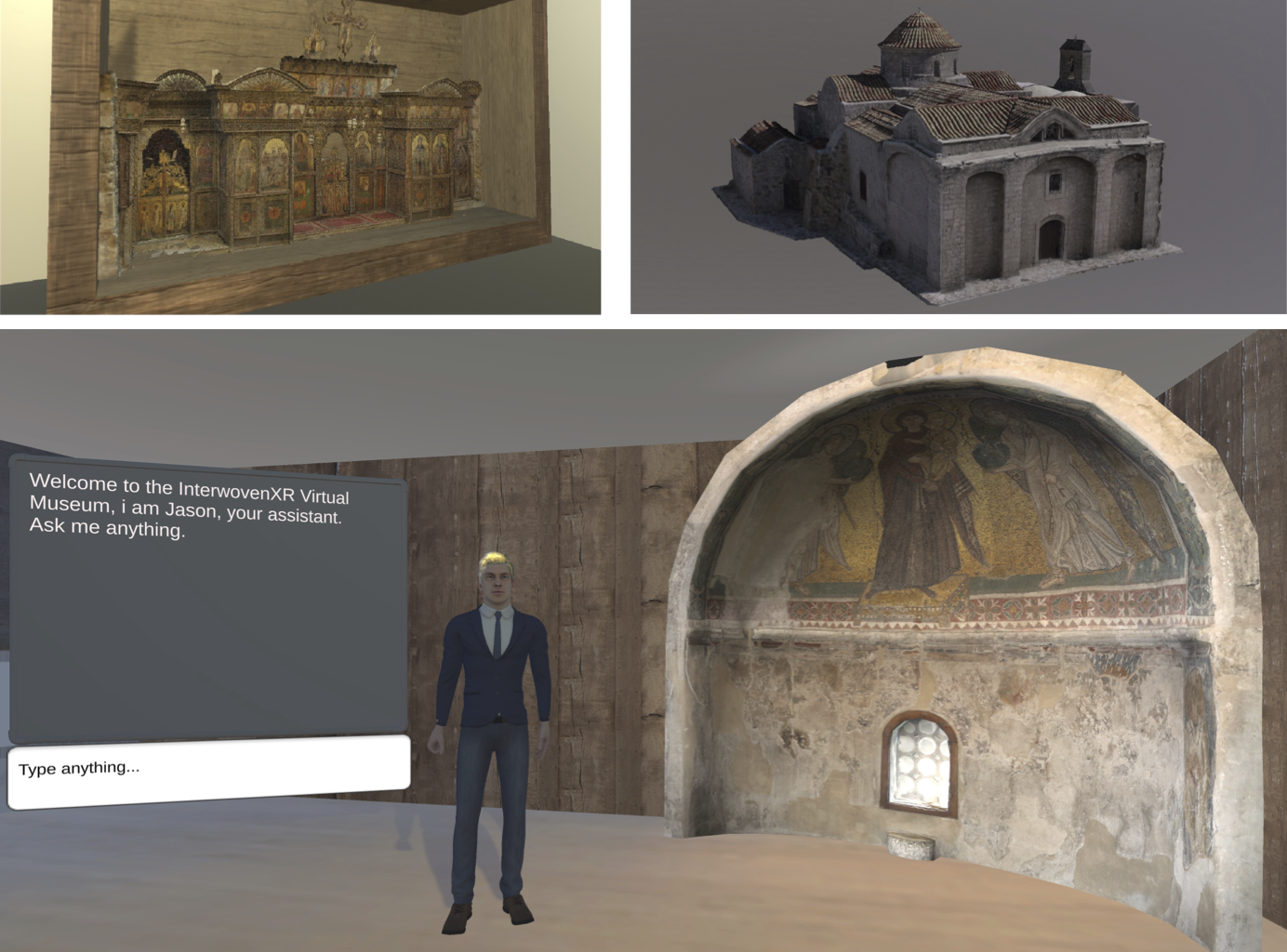}} 
    \caption{\textit{Example screenshots from the InterwovenXR testbed.}}\vspace*{-5pt} 
    \label{fig:CPSS_example_screenshots} 
\end{figure}

The virtual guide provide the primary interaction point at which AI service orchestration becomes necessary. 
A visitor may ask for museum orientation, request a translation, ask a factual question, request specialist interpretation, or ask the system to generate a 3D object for the virtual environment. 
From the user's perspective these interactions share the same channel, but each corresponds to a different computational task requiring a different backend service. 
This range of request types makes the virtual museum scenario a meaningful testbed for evaluating the proposed gateway, as it requires discrimination between semantically distant routes (e.g.\ translation vs.\ 3D generation) and semantically adjacent ones (e.g.\ general heritage queries vs.\ mosaic interpretation).

\subsection{SLM-Based Agent Orchestration Gateway}

\subsubsection{Design Requirements}
The design requirements were derived through a scenario-driven analysis of the \textit{InterwovenXR} virtual museum deployment. 
The interaction scenarios were mapped to the backend capabilities required to support them and the resulting integration requirements were  consolidated around client-service decoupling, heterogeneous backend interoperability, semantic service selection, routing validation, distributed deployment, and interaction latency to meet the following needs:
1) the virtual world client communicates with a single endpoint rather than maintaining separate integrations per service; 
2) the system supports heterogeneous backend types, including OpenAI-compatible chat services, retrieval-augmented generation (RAG) services, plain HTTP services, JSON-based services, and 3D generation systems; 
3) service routing is determined by the semantic meaning of user requests rather than static rules; 
4) routing decisions are structured and validated before service invocation; 
5) the architecture supports services distributed across edge, and cloud resources; 
and 6) the routing stage operates with latency suitable for interactive use and conversation response in $>$2 seconds \cite{Maslych2025}.

\subsubsection{System Overview}
The gateway is implemented as a Python-based FastAPI service that exposes a single REST endpoint. 
The virtual world submits each user prompt via a POST request; the gateway processes it through a routing pipeline and returns a JSON response containing the generated answer, selected route, service metadata, latency information, and any generated asset references. 
Figure~\ref{fig:gateway_architecture} illustrates the high-level request flow from the virtual world through the intent recognition mechanism to the selected backend services, which may reside on servers within the edge network or on remote servers and cloud infrastructure.
In the current prototype, the Agent Orchestration mechanism runs on an NVIDIA Jetson Orin NX 8GB edge device. 
Example backend services used for experimentation include a RAG server for general heritage knowledge, a fine-tuned LLM server for specialist mosaic interpretation, and a text-to-3D generation service based on OpenAI's Shap-E model.

\subsubsection{Architectural Components}

\begin{figure}[htbp]
\centerline{\includegraphics[width=0.2\textwidth]{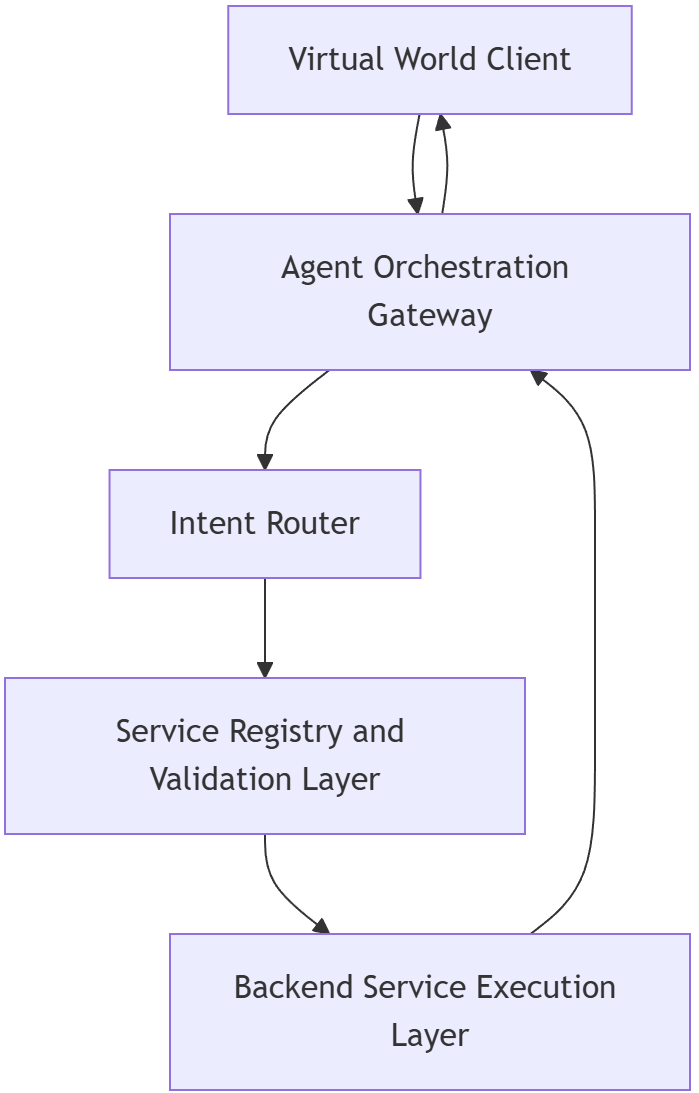}}
\caption{Orchestration Gateway runtime interaction flow overview.}
\label{fig:gateway_architecture}
\end{figure}

The architecture comprises five core components as shown in Fig.~\ref{fig:gateway_architecture}.

1) The \texttt{Virtual World Client} captures user input through an in-world interface attached to each virtual avatar (Fig \ref{fig:CPSS_example_screenshots}) serialises the prompt as JSON, and forwards it to the gateway. 
The client is decoupled from service selection with no knowledge of where individual services are hosted.

2) The \texttt{Agent Orchestration Gateway} is the main coordination component. 
It receives the request, forwards the prompt to (3) the \texttt{Intent Router}, validates returned routing decision, calls the selected service, and constructs the final response. 
It also abstracts the different protocols, payload formats, and output structures of the underlying services, providing the virtual world with a consistent interface.

\begin{figure}[htbp]
    \centerline{\includegraphics[width=20.5 pc]{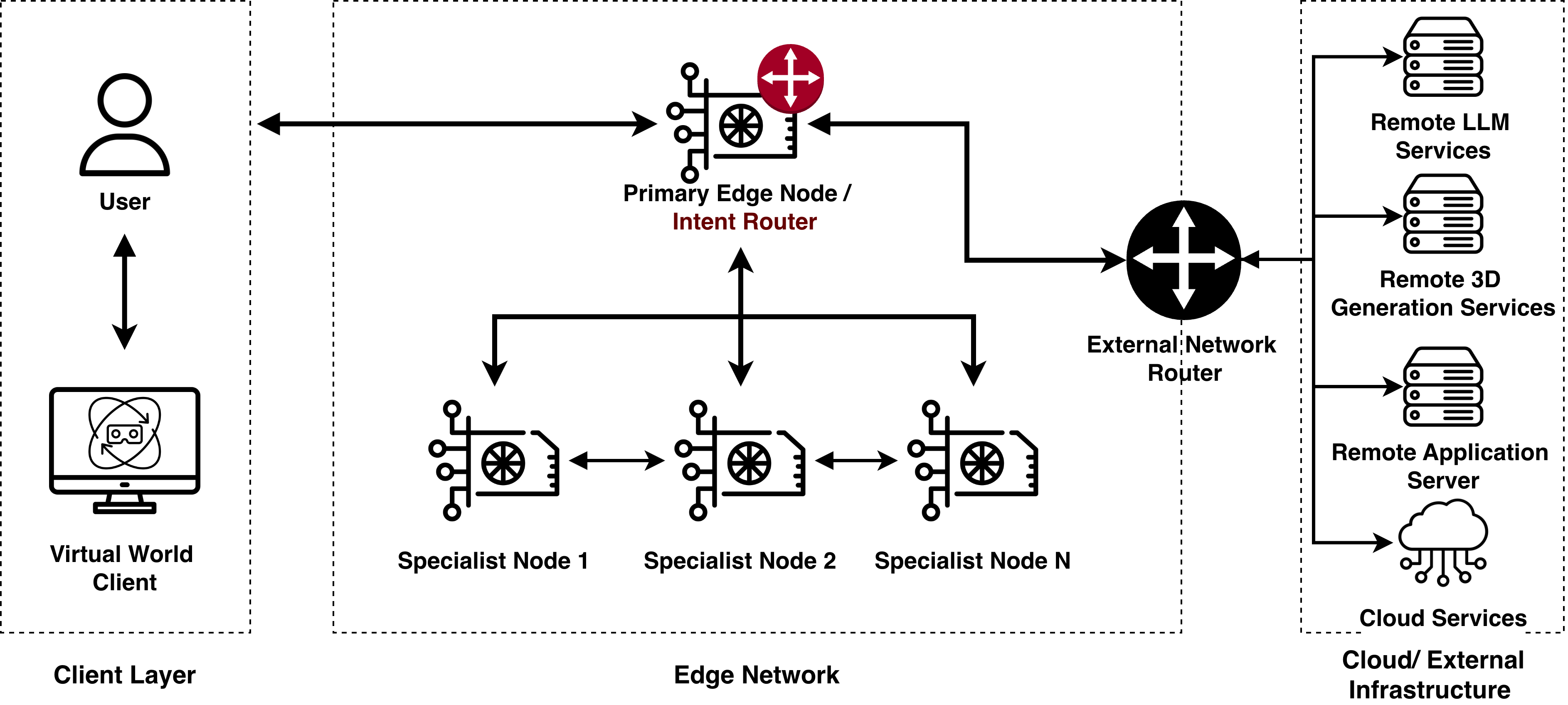}} 
    \caption{\textit{Example of the intent router within an edge/cloud architecture.}}\vspace*{-5pt} 
    \label{fig:edge_cloud_architecture} 
\end{figure}

3) The \texttt{Intent Router} performs intent classification using an SLM.
It receives a routing prompt generated dynamically from the service descriptions in the registry and returns a structured JSON decision with the selected route, intent label, confidence score, and a normalised query. 
For example, conversational requests are routed to a local response service on the same edge device; translation requests are forwarded to a language service within the edge network; factual heritage questions are directed to a RAG model on a remote server; and asset-generation requests are dispatched to a cloud-hosted 3D generation service (See Figure \ref{fig:edge_cloud_architecture}).

4) The \texttt{Service Registry and Validation} layer maintains the catalogue of available services, including their route names, types, endpoints, enabled status, and descriptions. 
It serves two functions: constructing the routing options provided to the \texttt{Intent Router}, and validating the router's output before execution. 
If the router returns an unrecognised or disabled route, the gateway applies a fallback route or requests clarification from the user.

5) The \texttt{Backend Service Execution} layer invokes the selected service according to the communication pattern defined in the registry. 
The current prototype supports OpenAI-compatible chat services, plain-text HTTP services, and JSON-based HTTP services.

\subsubsection{Runtime Flow}
At runtime (Fig.~\ref{fig:gateway_architecture}), the user submits a natural language prompt through the virtual agent interface. 
The virtual world client forwards the request to the gateway, which passes the prompt to the Intent Router for classification. 
The router returns the selected route, intent label, confidence score, and adapted query; the gateway validates this decision against the Service Registry, invokes the appropriate backend service, normalises the response into a structured JSON object, and returns it to the virtual world client for rendering.

\section{Research Methodology}
\label{sec:evaluation_methodology}
To investigate the efficacy of the SLM-based orchestration gateway, an evaluation study was conducted to examine whether SLMs can serve as intent-recognition and service-routing mechanisms for AI-driven virtual worlds, with particular attention to routing accuracy, output reliability, and latency under edge deployment constraints.
The evaluation was conducted within the \textit{InterwovenXR} virtual museum scenario, centred on a dedicated room focused on the Panagia Aggeloktisti church where visitors interact with virtual guide agents to request general guidance, translation, factual information about the church, specialist interpretation of the apse mosaic, or 3D content generation. 
This scenario was selected because it demands discrimination between semantically distant and also adjacent ones, making it a meaningful testbed for intent-routing evaluation.
To conduct the evaluation, two research questions have been devised:

\textit{• RQ1: To what extent can compact SLMs serve as intent-recognition mechanisms for service routing in AI-driven virtual worlds?} This question is addressed by evaluating multiple SLMs on a route-only task, comparing routing accuracy, route-specific performance, invalid-output rate, and latency on edge hardware.

\textit{• RQ2: Can a layered SLM configuration provide a more practical balance between intent recognition latency and local conversational response generation than a single-model setup?} 
This question examines whether assigning routing and response generation to separate compact models through a very small model for first-stage intent recognition and a larger compact model for conversational reply offers practical advantages over a single model handling both tasks.

The evaluation is structured accordingly into two stages. 
Stage~1 isolates the Intent Router from all downstream services to assess routing accuracy, route-specific behaviour, invalid-output rate, and latency independently of retrieval pipelines, remote inference, or 3D generation backends. 
Stage~2 evaluates a layered edge configuration, measuring end-to-end latency across the full interaction path for conversational prompts to assess suitability for real-time interaction.

\subsection{Routing Task}
The routing task is formulated as a five-class classification problem. 
Given a natural language visitor prompt, the Intent Router must select one of the following service routes:

\begin{itemize} 
    \item \textit{Conversational Route}: greetings, guidance, clarification, navigation, identity queries, and lightweight replies. 
    \item \textit{Language Adaptation Route}: translation and multilingual support. 
    \item \textit{Domain Expert Route}: specialist queries concerning the Panagia Aggeloktisti apse mosaic, including its artistic style, dating, and preservation. 
    \item \textit{Knowledge Retrieval Route}: general factual queries about the church via RAG, covering history, architecture, heritage value, religious context, and visitor interpretation. 
    \item \textit{Asset Generation Route}: 3D model generation in GLB format and dynamic loading into the virtual world. \end{itemize}

\begin{figure}[!h]
    \centering
    \includegraphics[width=1.0\linewidth]{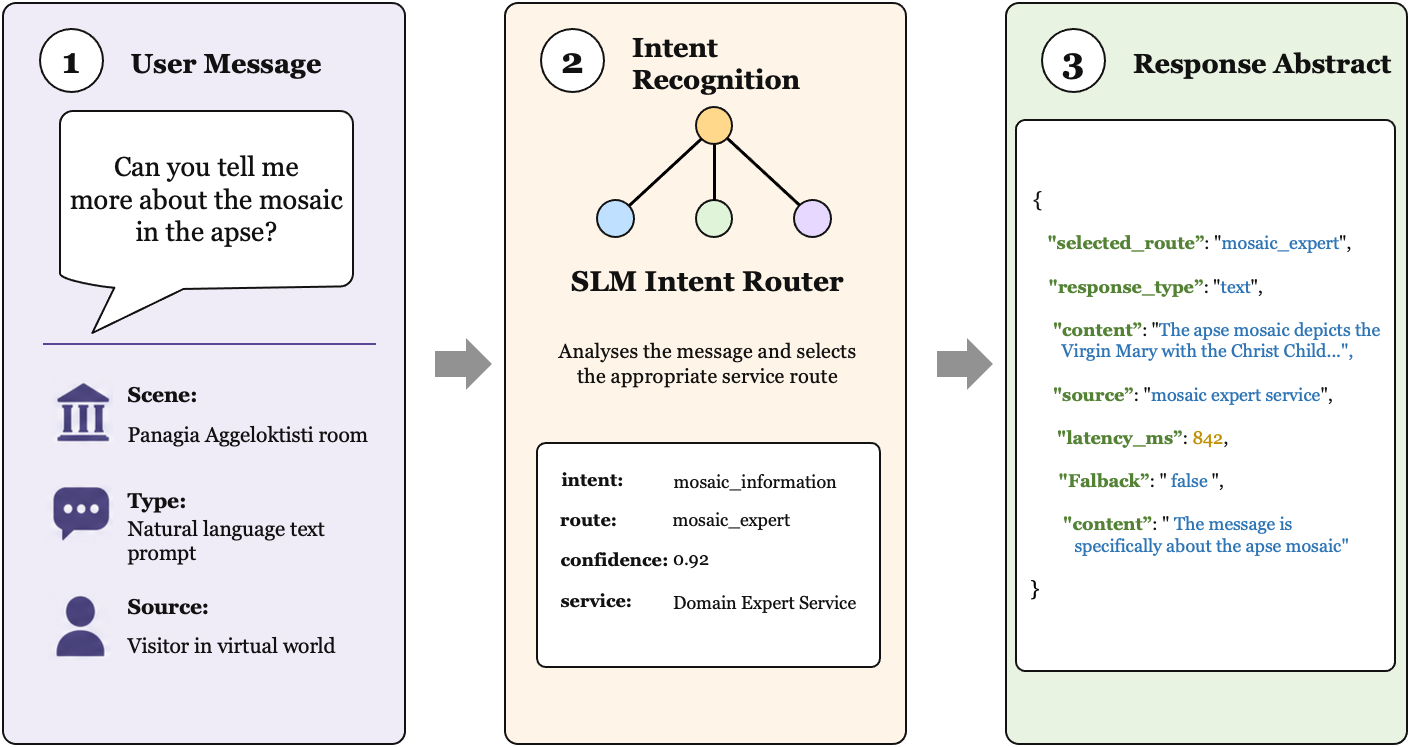}
    \caption{Abstract routing example in the Agent Orchestration Gateway.}
    \label{fig:gateway_overview}
\end{figure}

The router returns a structured JSON decision comprising the selected route, intent label, confidence score, and a brief routing rationale (Fig. \ref{fig:gateway_overview}). 
Backend services are not invoked in Stage~1, ensuring that measured performance reflects only the router's intent-recognition and output-formatting behaviour.
\subsection{Evaluation Dataset}
A dataset of 500 prompts was constructed for Stage~1, balanced across the five route classes at 100 prompts per class. 
The evaluation dataset combined author-written and synthetically generated prompts using ChatGPT 5.5, with ground-truth labels assigned according to the predefined service-route definitions and reviewed and validated by the authoring team.
Prompts were designed to reflect realistic visitor interactions, encompassing first-visit questions, translation requests, factual church queries, mosaic-specific interpretative questions, and 3D generation requests. 
Importantly, the dataset was constructed to require semantic understanding rather than keyword matching, for instance, prompts may reference the apse mosaic without using the word ``mosaic'', or frame factual questions in ways that superficially resemble specialist interpretation queries to reflect the natural language variability expected during interaction with virtual agents.
To validate dataset coherence and route separability, a Qwen2.5-14B-Instruct-Q5 model was used as a high-capacity reference. 
This model was not considered a candidate edge router; rather, its performance established an upper-bound indication of whether route categories were sufficiently distinguishable, helping to attribute lower accuracy in smaller models to model limitations rather than dataset ambiguity.

For Stage~2, 100 prompts assigned to the Conversational Route were submitted directly from the virtual world client. 
Unlike Stage~1, this phase measured the complete end-to-end interaction path, from prompt submission through the virtual world client, gateway routing, response generation on the local model, and response delivery back to the client.

\subsection{Evaluated SLM Models}
A wide set of model families were explored during preliminary testing to identify candidates capable of producing valid structured JSON and correct routing decisions. 
Models that failed to achieve this for more than 75\% of routing decisions and more than 25\% of valid JSON were excluded. 
The results of \textit{TinyLlama-1.1B-Chat-v1.0-Q4} and \textit{StableLM2-1.6B-Q4} as shown in Table \ref{tab:router_combined_results} are indicative failed examples.
From the preliminary exploration, \textit{Qwen2.5-1.5B-Instruct-Q4} and \textit{SmolLM2-1.7B-Instruct-Q4} were identified as the strongest-performing compact models and were selected as the primary Stage~1 routers. 
Their smaller variants (\textit{Qwen2.5-0.5B-Q4}, \textit{SmolLM2-360M-Q4}), were identified as sub-billion-parameter candidates for Stage~2, where low-latency intent recognition is prioritised and have been further adapted through fine-tuning.

\subsubsection{Fine-Tuning Sub-Billion Parameter Models}
Given the known limitations of sub-billion-parameter models in general instruction following, \textit{SmolLM2-360M} and \textit{Qwen2.5-0.5B} were fine-tuned on the specific routing task. 
Both were trained using supervised instruction tuning on a dataset of 5,000 routing examples (1,000 per route class), where each instance comprised a system instruction, a user query, and a ground-truth JSON routing response consistent with the Intent Router output format. 
Fine-tuning was performed using 4-bit QLoRA on a Windows workstation with an NVIDIA GeForce RTX 4060 GPU, with a per-device batch size of 1, gradient accumulation over 8 steps, across 3 epochs (1,689 optimisation steps). 
\textit{SmolLM2-360M-FT} (FT: Fine-Tuned version) achieved a validation loss of 0.0775 and mean token accuracy of 0.9740; \textit{Qwen2.5-0.5B-FT} achieved a validation loss of 0.0724 and mean token accuracy of 0.9745. 
Results suggest that the fine-tuned models learned to reproduce the expected JSON routing structure with strong consistency across the validation set. 

\subsubsection{Deployment Configuration}
The orchestrator gateway mechanism and candidate SLMs were deployed on an NVIDIA Jetson Orin NX (8GB RAM) using \textit{llama.cpp}, with model-layer offloading to CUDA. 
\textit{Llama.cpp} was selected for its lightweight footprint, quantised model support, and portability across heterogeneous hardware, consistent with the requirement for a routing layer that is independent of cloud infrastructure or specific accelerator types. 
The virtual world client and the Jetson were on the same local network to simulate the intended low-overhead edge deployment. 
The \textit{Qwen2.5-14B-Instruct-Q5} model was deployed separately on a desktop PC, as it exceeds the resource constraints of the edge configuration and was ony used for dataset validation.

\subsubsection{Evaluation Metrics}
Three primary metrics were recorded. 
\textit{Routing accuracy} is the percentage of prompts for which the router's selected route matched the ground-truth label; route-specific accuracy was additionally computed to identify performance variation across semantically distinct and adjacent categories. 
\textit{Routing latency} was measured in milliseconds from prompt submission to receipt of the routing response, capturing the overhead introduced at the gateway layer before any backend service is invoked. 
\textit{Invalid-output rate} is the proportion of responses that could not be parsed as valid routing decisions or mapped to a registered service, reflecting the router's instruction-following reliability under structured-output constraints.

\subsection{Experimental Procedure}
\label{lab:experimental_procedure}
\textit{Stage~1 (Routing Evaluation)}: The candidate SLMs, including the fine-tuned variants and the reference model, were evaluated against the full 500-prompt dataset. 
A custom script submitted each prompt to the Intent Router and recorded the selected route, intent label, confidence score, routing rationale, latency, and output validity. 
The routing prompt was generated dynamically from the service registry, mirroring the gateway's runtime behaviour. 
Backend services were not invoked, isolating routing behaviour and enabling direct measurement of output validity, since the gateway must parse and validate the routing decision prior to service dispatch. 
\textit{Stage~2 (Layered Configuration Evaluation)}: The fine-tuned sub-billion-parameter model served as the first-stage intent router, with the larger SLM handling conversational response generation. 
100  prompts were submitted directly from the virtual world environment, and end-to-end latency was recorded from prompt dispatch to response receipt in the client.

\section{Results}
\subsection{SLM Intent Recognition and Service Routing Evaluation}
\label{sec:results}

\begin{table*}[!bh]
\centering
\caption{Overall intent-recognition mechanism evaluation results.}
\label{tab:router_combined_results}
\renewcommand{\arraystretch}{1.15}
\resizebox{\textwidth}{!}{%
\begin{tabular}{lccccccccc}
\toprule
\textbf{Router model} 
& \textbf{Correct} 
& \textbf{Accuracy} 
& \makecell{\textbf{Average Routing}\\\textbf{Latency}} 
& \textbf{Conversation} 
& \textbf{Language} 
& \makecell{\textbf{Mosaic}\\\textbf{Expert}} 
& \makecell{\textbf{Heritage}\\\textbf{Knowledge}} 
& \makecell{\textbf{3D Generation}\\\textbf{Service}} 
& \textbf{Invalid} \\
\midrule

\multicolumn{10}{c}{\textbf{Qwen 2.5 Models}} \\
\midrule
Qwen2.5-0.5B    
& 124/500 & 24.80\% & 329.3 ms 
& 37.00\% & 0.00\% & 69.00\% & 0.00\% & 18.00\% & 0.00\% \\

Qwen2.5-0.5B-FT    
& 415/500 & 83.00\% & 392.0 ms 
& 93.00\% & 100.00\% & 49.00\% & 89.00\% & 84.00\% & 3.20\% \\

Qwen2.5-1.5B    
& 473/500 & 94.60\% & 1776.9 ms 
& 91.00\% & 100.00\% & 87.00\% & 95.00\% & 100.00\% & 0.00\% \\

\midrule
\multicolumn{10}{c}{\textbf{SmolLM2 Models}} \\
\midrule
SmolLM2-360M    
& 84/500 & 16.80\% & 533.5 ms 
& 78.00\% & 0.00\% & 6.00\% & 0.00\% & 0.00\% & 9.00\% \\

SmolLM2-360M-FT    
& 246/500 & 49.20\% & 560.3 ms 
& 12.00\% & 89.00\% & 84.00\% & 0.00\% & 61.00\% & 22.60\% \\

SmolLM2-1.7B    
& 427/500 & 85.40\% & 1314.9 ms 
& 92.00\% & 97.00\% & 78.00\% & 60.00\% & 100.00\% & 0.80\% \\

\midrule
\multicolumn{10}{c}{\textbf{Indicative Models that Failed Routing and Validity Requirements}} \\
\midrule
StableLM2-1.6B    
& 177/500 & 35.40\% & 1972.6 ms 
& 93.00\% & 24.00\% & 0.00\% & 6.00\% & 54.00\% & 10.60\% \\

TinyLlama-1.1B    
& 60/500 & 12.00\% & 1654.0 ms 
& 56.00\% & 4.00\% & 0.00\% & 0.00\% & 0.00\% & 38.40\% \\

\midrule
\multicolumn{10}{c}{\textbf{Reference Model for Dataset Validation}} \\
\midrule
Qwen2.5-14B    
& 489/500 & 97.80\% & 5684.1 ms 
& 100.00\% & 99.01\% & 100.00\% & 95.24\% & 95.24\% & 0.00\% \\

\bottomrule
\end{tabular}%
}
\end{table*}
The evaluation results are shown in Table \ref{tab:router_combined_results}. 
\textit{Qwen2.5-14B} achieved 97.80\% overall routing accuracy with no invalid outputs, validating the five routing categories used in the evaluation suggesting that they were sufficiently distinguishable when model capacity was not the main limiting factor, providing confidence that the dataset and routing schema were coherent for the comparative evaluation of smaller models. 

From the high parameter SLM versions, \textit{Qwen2.5-1.5B} achieved the strongest routing performance, scoring an overall accuracy of 94.60\%, with no invalid outputs, and consistently correct across all service routes, indicating strong instruction-following behaviour and reliable identification between the heterogeneous service intents.
\textit{SmolLM2-1.7B} demonstrated a good level of routing performance, reaching 85.40\% overall accuracy but also lower average routing latency of 1314.9 ms. 
Its performance was strong for conversational prompts, language adaptation, mosaic expert requests, and 3D generation requests, but it was weaker for the Heritage Knowledge Route, where it achieved 60.00\%. 
This suggests that although \textit{SmolLM2-1.7B} can follow the structured routing task in many cases, it was less reliable in distinguishing general heritage knowledge from adjacent interpretative categories. 
\textit{StableLM2-1.6B} and \textit{TinyLlama-1.1B} are used as examples of limited suitability for the proposed routing mechanism, achieving low accuracy and high rate of invalid outputs.

From the sub-billion parameter versions, \textit{Qwen2.5-0.5B}, achieved the lowest average routing latency at 329.3 ms, but its overall accuracy was only 24.80\%. 
\textit{SmolLM2-360M} also demonstrated low latency at 533.5 ms, but achieved the lower overall accuracy with only 16.80\% routing accuracy and 9.00\% invalid outputs.
These results confirm our initial assumption that these base models on the sub-billion parameters are not able to perform reliable multi-service routing by default without task-specific adaptation, despite its strong latency advantage.  
The results of their corresponding fine-tuned versions show substantial improvements in routing reliability, without significantly compromising latency.
Most notably, \textit{Qwen2.5-0.5B-FT} has substantially improved in routing accuracy, achieving 83.00\% routing accuracy over the 24.80\% accuracy of the base model, corresponding to a 58.2\% improvement. 
Its average routing latency increased only slightly, from 329.3 ms to 392.0 ms, which indicates that task-specific adaptation preserved the main latency advantage of the smaller model. 
It is noted however that the fine-tuned model still underperformed compared with \textit{Qwen2.5-1.5B}, particularly in cases requiring fine-grained semantic separation between related routes, and this opens the scope for further future fine tuning refinements and evaluation.
The fine-tuned version of \textit{SmolLM2-360M-FT} also improved over its base model, increasing routing accuracy from 16.80\% to 49.20\%, corresponding to a 32.4\% improvement. 
Its average routing latency increased slightly, from 533.5 ms to 560.6 ms. 
However, the model still produced a relatively high invalid output rate of 22.60\% and failed to reliably distinguish some service categories, particularly the heritage knowledge route. 
This suggests that, although fine-tuning improved the model’s routing behaviour, \textit{SmolLM2-360M-FT} was less reliable than the \textit{Qwen2.5-0.5B-FT} model for this task.

\subsection{Layered Routing and Conversational Configuration}

\begin{figure*}[!bh]
    \centering
    \includegraphics[width=1\linewidth]{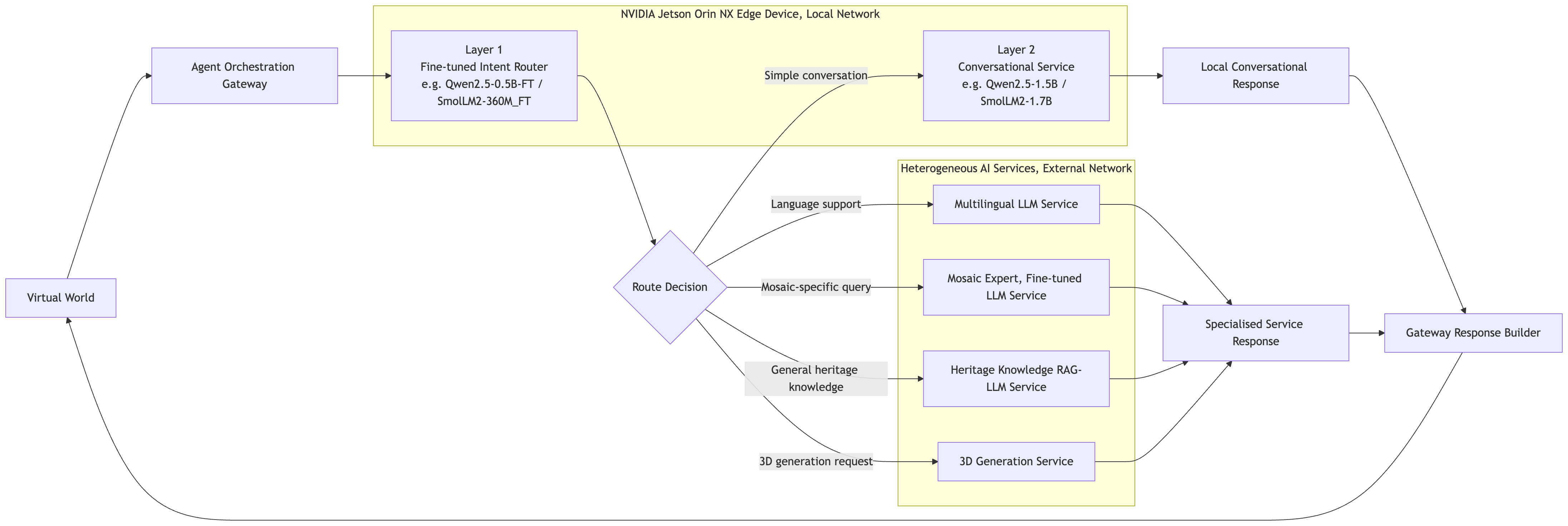}
    \caption{Multi-layer agent orchestration architecture overview. A fine-tuned compact router performs the first-stage intent classification at the edge, and simple conversational handling and specialised AI services are invoked according to the selected route.}
\label{fig:multilayer_agent_orchestration}
\end{figure*}

\begin{table*}[!tb]
\centering
\caption{End-to-end latency results for combined intent-router and conversational-route configurations.}
\label{tab:end_to_end_latency_combinations}
\renewcommand{\arraystretch}{1.15}
\resizebox{\textwidth}{!}{%
\begin{tabular}{llccccccc}
\toprule
\makecell{\textbf{Intent Recognition}\\\textbf{Router Model}} 
& \makecell{\textbf{Conversational}\\\textbf{Route Model}} 
& \makecell{\textbf{Average}\\\textbf{Latency}} & \textbf{Median} & \textbf{Min} & \textbf{Max} & \textbf{P90} & \textbf{P95} & \textbf{P99} \\
\midrule

Qwen2.5-0.5B-FT & Qwen2.5-1.5B 
& 1454.18 ms & 1415.93 ms & 999.17 ms & 3350.69 ms & 1734.49 ms & 1995.68 ms & 2640.59 ms \\

Qwen2.5-0.5B-FT & SmolLM2-1.7B 
& 1518.63 ms & 1358.18 ms & 1049.41 ms & 3715.38 ms & 2050.79 ms & 2174.89 ms & 2644.32 ms \\

SmolLM2-360M-FT & Qwen2.5-1.5B 
& 1306.06 ms & 1247.34 ms & 865.93 ms & 2816.36 ms & 1836.50 ms & 2116.62 ms & 2312.07 ms \\

SmolLM2-360M-FT & SmolLM2-1.7B 
& 1339.26 ms & 1265.26 ms & 863.53 ms & 4015.71 ms & 1902.80 ms & 2071.19 ms & 2646.67 ms \\

\bottomrule
\end{tabular}%
}
\end{table*}
Stage 2 evaluated whether intent recognition and conversational response generation could be separated across two compact models co-deployed on the same edge device, combining the low-latency routing behaviour of a fine-tuned sub-billion-parameter model with the stronger generative capability of a larger compact SLM for first-level conversational responses (See Figure  \ref{fig:multilayer_agent_orchestration}). 
In each configuration, the fine-tuned router handled the routing decision and, upon classification as a conversational prompt, the larger model generated the response locally without invoking a remote service.

Four pairings were evaluated as shown in Table \ref{tab:end_to_end_latency_combinations}: 1) \textit{Qwen2.5-0.5B-FT} with \textit{Qwen2.5-1.5B}, 2) \textit{Qwen2.5-0.5B-FT} with \textit{SmolLM2-1.7B}, 3) \textit{SmolLM2-360M-FT} with \textit{Qwen2.5-1.5B}, and 4) \textit{SmolLM2-360M-FT} with \textit{SmolLM2-1.7B}. 
In each pairing, both models were loaded simultaneously on the NVIDIA Jetson Orin NX and deployed through the \textit{llama.cpp} server a with memory usage  ranging from approximately between 5~GB at load, to 7.4~GB RAM during inference. 
This confirm that such pairing layered configuration can be deployable within the constraints of low- to mid-range edge hardware.
End-to-end latency was measured across 100 conversational prompts submitted directly from the virtual world client, covering the complete path from prompt dispatch through gateway routing and local response generation to delivery back to the client.

All four configurations achieved average end-to-end latencies below 1.6~s, with median values below 1.5~s (Table~\ref{tab:end_to_end_latency_combinations}), demonstrating low system latency for the evaluated conversational pipeline. 
This is below the 2 second threshold as suggested by \cite{Maslych2025} for effective interactive use.
The two configurations using \textit{SmolLM2-360M-FT} as the router achieved the lowest average latencies overall.
\textit{SmolLM2-360M-FT} paired with \textit{Qwen2.5-1.5B} reached 1306.06~ms average (median 1247.34~ms), and paired with \textit{SmolLM2-1.7B} reached 1339.26~ms (median 1265.26~ms). 
The \textit{Qwen2.5-0.5B-FT} pairings were slightly higher, at 1454.18~ms with Qwen2.5-1.5B and 1518.63~ms with \textit{SmolLM2-1.7B}, reflecting the higher base routing latency of the 0.5B model relative to the 360M model but with better routing accuracy.
Most notably, the results show that the layered configuration beyond being feasible on the edge device, it was also faster than using the larger compact model as the primary intent-recognition mechanism. 
This key finding suggests that separating intent recognition from conversational generation can reduce the computational cost of the first decision stage, but still preserve the ability to produce coherent  responses when the prompt is routed to the conversational layer. 

Considering the routing accuracy results from Stage~1 and the end-to-end latency results from Stage~2, the \textit{Qwen2.5} family pair offers the most balanced profile in our configuration. 
Although the \textit{SmolLM2-360M-FT} router produced lower latency in Stage 2, the \textit{Qwen2.5-0.5B-FT} router achieved stronger routing reliability in Stage 1, reaching 83.00\% routing accuracy and maintained a combined average end-to-end latency of 1454.18~ms for both models loaded together. 
This result indicates that the additional routing cost of the 0.5B model remains low when considered within the full interaction pipeline, particularly given the improved reliability of the routing decision.
\textit{Qwen2.5} family pairing is therefore the strongest candidate, combining accurate intent recognition with locally generated conversational responses within the interaction range suitable for AI-driven virtual worlds. 
The overall findings support the use of a layered two-model orchestration strategy for edge-based AI services in virtual worlds, demonstrating that the configuration of a fine-tuned compact router for fast service selection that activates a larger conversational model only when needed is more efficient than assigning routing and response generation to a single SLM. 

\section{Discussion}
The evaluation shows that compact SLMs can support practical intent recognition and service routing in AI-driven virtual worlds, addressing RQ1. 
Base sub-billion-parameter models were unreliable without adaptation, but task-specific fine-tuning substantially improved routing performance, particularly for Qwen2.5-0.5B, while retaining low latency. 
This is a practically significant finding as it suggests that very small models, which would otherwise be dismissed as unsuitable for multi-service routing, can be adapted into reliable edge routers through targeted supervised tuning.
The results also highlight the importance of route-boundary design. 
Errors were more likely when prompts belonged to semantically adjacent categories, indicating that the design of an AI service registry is a complex combination of software-engineering and modelling tasks influencing router performance and its ability to reliably distinguish between service intents. 
For virtual-world developers, this means that effective AI orchestration requires careful definition of service responsibilities rather than simply adding more backend models.

The layered configuration addresses RQ2 and provides the strongest systems-level contribution of the study. 
Co-deploying a fine-tuned sub-billion-parameter router with a larger compact conversational model on the device demonstrates that edge hardware can support both service selection and local conversational response generation. 
More importantly, the layered configuration was faster than using the larger compact model as the primary intent-recognition mechanism. 
This indicates that separating routing from response generation can reduce the computational cost of the first decision stage and preserve the ability to generate higher-quality local replies when the conversational route is selected.
The conversational latency results fall within a suitable range for first-level virtual-world dialogue with virtuak agents \cite{Maslych2025}, strengthening the argument that edge-deployed SLM orchestration can support responsive interaction in virtual environments.
This also suggest that for longer backend calls (e.g 3D generation), the approach may require feedback, progress indicators, or conversational fillers to preserve user engagement.

Overall, the findings show that SLM-based orchestration can make AI-driven virtual worlds more scalable and extensible. 
The gateway allows virtual agents to act as unified access points to distributed AI services, with the client remaining decoupled from model endpoints, service formats, and deployment locations. 
This has direct implications for immersive environments requiring multiple AI capabilities and the main contribution is therefore the demonstration of an architectural pattern in which compact edge-deployed models can coordinate heterogeneous generative AI services in real time.

\section{Conclusion, Limitations and Future Work}
This paper presented an SLM-based Agent Orchestration Gateway for AI-driven virtual worlds. 
The architecture provides a runtime coordination layer that routes natural-language requests to heterogeneous AI services through an edge-deployed intent-recognition mechanism, allowing services to be added or redistributed without modifying the client.
The evaluation showed that compact SLMs can support intent-based routing and that fine-tuning can make sub-billion-parameter models viable for low-latency orchestration. 
The layered configuration further demonstrated the feasibility of separating service selection from conversational response generation on edge hardware.

The study has several limitations. 
Evaluation was conducted using a controlled 500-prompt dataset within a single virtual museum scenario, limiting generalisability to other domains and interaction styles. 
Response quality and user experience were not evaluated, and the fine-tuned models remain specific to the defined routing schema and service set. 
The fine-tuned models were trained for a specific routing schema and service set, reducing transferability without retraining, and fine-tuning introduces maintenance overhead when route definitions or service configurations evolve.
Performance was also assessed on a single edge device under controlled network conditions.

Future work will evaluate additional SLMs and quantisation strategies, larger and more diverse training datasets, multi-turn and ambiguous prompts, and additional virtual-world scenarios. 
Future user studies will assess perceived responsiveness, conversational quality, task success, engagement, immersion, and acceptance. 
Further development will also investigate improved fallback mechanisms, response validation, and adaptive model-selection policies balancing accuracy, latency, device load, and service availability across heterogeneous deployment environments.

\textbf{Data and Reproducibility:} The evaluation and fine-tuning datasets used in this study are publicly available\footnote{https://github.com/rays-and-vectors/InterwovenXR/tree/main/IntentRouting}.
\bibliographystyle{ieeetr}
\bibliography{References}

\end{document}